\documentclass[sigconf]{acmart}
\usepackage{xcolor}
\usepackage{float}
\usepackage{booktabs}

\definecolor{darkred}{RGB}{192, 0, 0}

\AtBeginDocument{%
  }

\copyrightyear{2025}
\acmYear{2025}
\setcopyright{acmlicensed}
\acmConference[MM '25] {Proceedings of the 33rd ACM International Conference on Multimedia}{October 27--31, 2025}{Dublin, Ireland.}
\acmBooktitle{Proceedings of the 33rd ACM International Conference on Multimedia (MM '25), October 27--31, 2025, Dublin, Ireland}
\acmISBN{979-8-4007-2035-2/2025/10}
\acmDOI{10.1145/3746027.3758269}

\begin{document}

\title{UniSVG: A Unified Dataset for Vector Graphic Understanding and Generation with Multimodal Large Language Models}

\author{Jinke Li}
\authornote{Work done during internship at Tencent.}
\authornote{Equal contribution.}
\email{jinke.23@intl.zju.edu.cn}
\orcid{0009-0009-9426-9230}
\affiliation{%
  \institution{Zhejiang University}
  \city{Haining}
  \state{Zhejiang}
  \country{China}
}

\author{Jiarui Yu}
\authornotemark[2]
\email{yujiarui9910@gmail.com}
\orcid{0000-0001-6327-6811}
\affiliation{%
  \institution{Tencent}
  \city{Shenzhen}
  \state{Guangdong}
  \country{China}
}

\author{Chenxing Wei}
\authornotemark[1]
\email{weichenxing2023@email.szu.edu.cn}
\orcid{0009-0005-4545-596X}
\affiliation{%
  \institution{Shenzhen University}
  \city{Shenzhen}
  \state{Guangdong}
  \country{China}}
  
\author{Hande Dong}
\authornote{Corresponding authors.}
\email{donghd66@gmail.com}
\orcid{0000-0003-0074-2664}
\affiliation{%
 \institution{Tencent}
 \city{Shenzhen}
 \state{Guangdong}
 \country{China}}
 
\author{Qiang Lin}
\email{cheaterlin@tencent.com}
\orcid{0009-0001-2005-6170}
\affiliation{%
 \institution{Tencent}
 \city{Shenzhen}
 \state{Guangdong}
 \country{China}}

\author{Liangjing Yang}
\authornotemark[3]
\email{liangjingyang@intl.zju.edu.cn}
\orcid{0000-0002-3294-0879}
\affiliation{%
  \institution{Zhejiang University}
  \city{Haining}
  \state{Zhejiang}
  \country{China}}
  
\author{Zhicai Wang}
\email{zhicaiw@outlook.com}
\orcid{0000-0001-6781-3442}
\affiliation{%
  \institution{Independent Researcher}
  \city{Hefei}
  \state{Anhui}
  \country{China}}

\author{Yanbin Hao}
\email{haoyanbin@hotmail.com}
\orcid{0000-0002-0695-1566}
\affiliation{%
  \institution{Hefei University of Technology}
  \city{Hefei}
  \state{Anhui}
  \country{China}}

\renewcommand{\shortauthors}{Li et al.}

\begin{abstract}
Unlike bitmap images, scalable vector graphics (SVG) maintain quality when scaled, frequently employed in computer vision and artistic design in the representation of SVG code. In this era of proliferating AI-powered systems, enabling AI to understand and generate SVG has become increasingly urgent. However, AI-driven SVG understanding and generation (U\&G) remain significant challenges. SVG code, equivalent to a set of curves and lines controlled by floating-point parameters, demands high precision in SVG U\&G. Besides, SVG generation operates under diverse conditional constraints, including textual prompts and visual references, which requires powerful multi-modal processing for condition-to-SVG transformation. Recently, the rapid growth of Multi-modal Large Language Models (MLLMs) have demonstrated capabilities to process multi-modal inputs and generate complex vector controlling parameters, suggesting the potential to address SVG U\&G tasks within a unified model. To unlock MLLM’s capabilities in the SVG area,  we propose an SVG-centric dataset called UniSVG, comprising 525k data items, tailored for MLLM training and evaluation. To our best knowledge, it is the first comprehensive dataset designed for unified SVG generation (from textual prompts and images) and SVG understanding (color, category, usage, etc.). As expected, learning on the proposed dataset boosts open-source MLLMs' performance on various SVG U\&G tasks, surpassing SOTA close-source MLLMs like GPT-4V.
We release dataset, benchmark, weights, codes and experiment details on \href{https://ryanlijinke.github.io/}{https://ryanlijinke.github.io/}.

\end{abstract}

\ccsdesc[500]{Computing methodologies~Shape inference}
\ccsdesc[300]{Applied computing~Arts and humanities}

\keywords{Scalable vector graphics, Dataset, MLLM finetuning}

\begin{teaserfigure}
  \centering
  \includegraphics[width=1\textwidth]{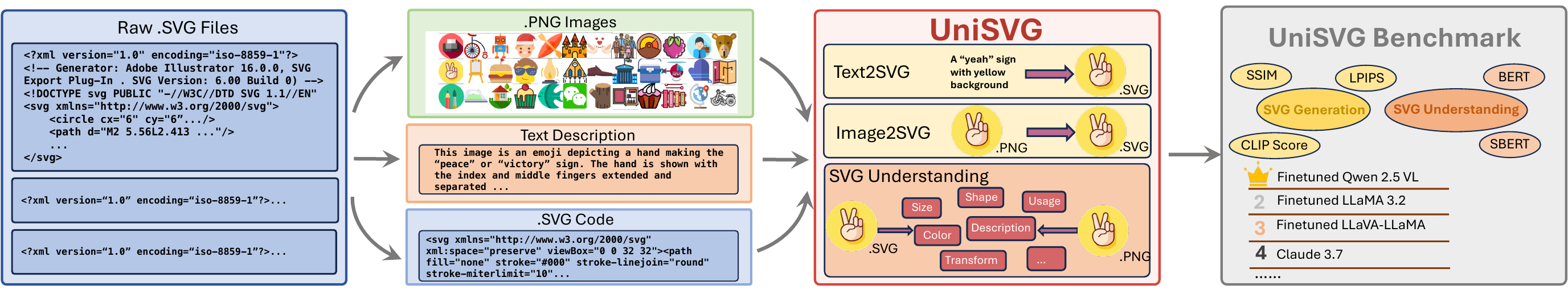}
  \caption{Overview of our proposed UniSVG dataset, task, and benchmark on MLLMs}
  \label{figure1}
\end{teaserfigure}

\maketitle
\section{Introduction}

Vector graphics are a type of computer graphics where images are created using geometric shapes like points, lines, curves, and polygons on a Cartesian plane, and can be resized infinitely without any loss of quality.  One of the most popular formats of vector graphics is Scalable Vector Graphics (SVGs)~\cite{quint2003scalable}, which are composed of XML-based code that defines the shapes, paths, colors, and text within the image, and are favored for their high-quality rendering, small file sizes and dynamic editability. Their versatility and precision make them widely used in computer vision and art design.

Despite the numerous benefits of SVG, enabling machines to understand and generate SVG remains a significant challenge. RNN-based approaches ~\cite{lopes2019learnedrepresentationscalablevector, Reddy_2021_CVPR}, layer-wise path optimization ~\cite{Ma_2022_CVPR} and diffusion-related methods ~\cite{Jain_2023_CVPR,xing2024svgdreamer} have achieved commendable results in SVG generation. However, their limited functionality, complex pipeline and costly post-processing limit their overall effectiveness. Moreover, in real-world applications, intelligent SVG-centric systems are expected to understand users' multi-modal inputs, including textual prompts and visual references.
To generate contextually appropriate SVGs based on diverse multi-modal conditions and achieve comprehensive interpretation, it is urgent to develop multi-modal AI-driven SVG generation and understanding.

Simultaneously, recent years have witnessed the rapid development of multi-modal large language models (MLLMs)~\cite{zhu2023minigpt, liu2023improvedllava, bai2023qwen, gpt4v, anthropic2023claude}, which demonstrate remarkable capabilities in processing complex cross-modal tasks~\cite{hu2024mplug, ning2025gns,gao2023g, ning2025gns,NEURIPS2024_d0822540, lee2024visual}.
This trajectory of improvement is expected to persist, as MLLM capabilities are enhanced by the continued scaling of their foundational LLM components, consistent with scaling laws~\cite{liu2024llavanext}.
Given their advanced cross-modal abilities and significant potential, we believe that fine-tuned MLLMs may serve as promising core of AI-driven systems dedicated to SVG understanding and generation tasks. To validate this hypothesis, we collect a dataset, enhance MLLMs through fine-tuning, benchmark them along with closed-source APIs, and finally demonstrate their capabilities through practical applications.

Several attempts have been made to integrate MLLMs into SVG-related tasks and benchmark their capabilities~\cite{nishina2024svgeditbench, zou2024vgbench, rodriguez2024starvector, yang2025omnisvgunifiedscalablevector}. 
As shown in Table \ref{table1}, researchers have designed various SVG tasks to investigate the efficacy of MLLMs in addressing SVG understanding and generation (U\&G) challenges.
These SVG tasks can be categorized into three types: conversion from scalar images to SVG code (Image2SVG), conversion from descriptive text to SVG code (Text2SVG), and SVG-related understanding tasks (SVG Understanding).
While benchmarks for these tasks have been established to assess MLLM performance, existing datasets often either lack training data~\cite{nishina2024svgeditbench, zou2024vgbench} or focus narrowly on one or two specific SVG U\&G tasks~\cite{rodriguez2024starvector}, proving insufficient for comprehensive development.
Consequently, there still exists a pressing need for a large-scale, comprehensive and open-source SVG-centric dataset specifically designed for both training and evaluation of MLLMs. 

To overcome the aforementioned limitations, we propose UniSVG, a comprehensive SVG-centric dataset with over 528k multi-modal data items for training MLLMs. After rigorous data cleaning and de-duplication, we obtained a substantial number of clean SVG samples suitable for MLLM training. Based on these SVG codes, we rendered images, generated detailed descriptions, and constructed relevant SVG question-answering pairs with GPT-4V ~\cite{gpt4v}. As shown in Figure \ref{figure1}, UniSVG leverages its multi-modal nature (visuals, text, and code), enabling it to encompass three primary tasks: image-to-SVG generation (ISVGEN), text-to-SVG generation (TSVGEN), and SVG understanding (SVGUN). We sampled 2,850 data items to create the UniSVG-benchmark, a test set for evaluating MLLMs on SVG-centric tasks. It includes diverse evaluation metrics to comprehensively assess model performance in SVG U$\&$G.

\renewcommand{\arraystretch}{1.2}

\begin{table}[!t]
\small
\belowrulesep=0pt
\aboverulesep=0pt
\centering
\caption{\textbf{Benchmark on MLLM solving SVG-centric tasks.}}
\vspace{-0.2cm}

\scalebox{0.7}{
\begin{tabular}{c|ccccc}
\toprule
\textbf{Dataset} & \textbf{Tasks} & \textbf{Train Set} & \textbf{Test Set} & \textbf{Multi Tasks} & \textbf{Open Source}\\
\midrule
SVGEditBench\cite{nishina2024svgeditbench} & SVG2SVG & --  & 100 & $\times$  & $\checkmark$ \\
\midrule
\multirow{2}{*}{VGBench\cite{zou2024vgbench}} & Image2SVG & \multirow{2}{*}{--}  & \multirow{2}{*}{10k} & \multirow{2}{*}{$\checkmark$}  & \multirow{2}{*}{$\checkmark$}\\
& SVG Understanding & & &  & \\
\midrule
SVG-Stack\cite{rodriguez2024starvector} & Image2SVG, Text2SVG & 4M & 13k & $\times$ & $\times$ \\
\midrule
\multirow{2}{*}{UniSVG (Ours)} & Image2SVG, Text2SVG & 525k & \multirow{2}{*}{3k} & \multirow{2}{*}{$\checkmark$} & \multirow{2}{*}{$\checkmark$} \\
& SVG Understanding & (2.2M) & & \\
\bottomrule
\end{tabular}
}
\label{table1}
\vspace{-0.2cm}
\end{table}
\renewcommand{\arraystretch}{1}

The UniSVG dataset facilitates the exploration of MLLMs' latent potential in addressing SVG-related tasks. Based on it, we finetuned multiple open-source MLLMs~\cite{liu2023improvedllava,bai2025qwen25vltechnicalreport,dubey2024llama}, and evaluated them against state-of-the-art models~\cite{gpt4v, anthropic2023claude}. Experiments show that fine-tuned MLLMs significantly improve in handling SVG tasks, even surpassing leading closed-source models. This indicates that the UniSVG dataset effectively unlocks MLLM potential, making them suitable for efficient SVG processing and manipulation in AI systems.

To summarize, our key contributions are as follows:
\begin{itemize}
    \item We propose UniSVG, the first large-scale, multi-task, open-source SVG-centric dataset for unified generation and understanding, supporting MLLM training and evaluation.
    \item We introduce UniSVG benchmark, along with diverse evaluation metrics, designed to evaluate the SVG generation and understanding capabilities. 
    \item Through our experiments and demo demonstration, we reveal the potential for advanced open-source MLLM-based systems in efficient SVG processing and manipulation. 
\end{itemize}

\section{Related Work}
\label{sec:related work}

\subsection{Deep Vector Graphics Generation} 

Deep vector graphics generation proposes addressing SVG generation challenges using deep learning. As a pioneering model, SketchRNN~\cite{ha2017neuralrepresentationsketchdrawings} employed an RNN-based encoder-decoder to reconstruct vector graphics from human-drawn sketches. Built on this, Lopes et al.~\cite{lopes2019learnedrepresentationscalablevector} developed a method to generate font characters in simplified SVG forms, such as straight lines and Bézier curves. Carlier et al.~\cite{carlier2020deepsvghierarchicalgenerativenetwork} further explored training transformer-based models to predict shapes in SVGs. While Wu et al.~\cite{10.1145/3618364} formulate SVG paths generation from text as sequence modeling with transformers. However, these methods struggle with producing more complex SVGs. Thus, some works focus on directly learning curve parameters. Researchers have achieved text-to-curve with the assistance of CLIP judgment~\cite{frans2021clipdrawexploringtexttodrawingsynthesis} or pre-trained text-to-image diffusion model~\cite{jain2022vectorfusiontexttosvgabstractingpixelbased}, and also have developed image-to-curve conversion through specially designed loss~\cite{Ma_2022_CVPR}. Although these methods have achieved acceptable results, their poor instruction following, limited functionality, and excessive inference iterations reduce practicality. Thus, developing a multimodal-aware model for end-to-end vector graph generation and understanding is essential.

\subsection{Multimodal Large Language Model(MLLM)} 

With the development of Large Language Models(LLMs)~\cite{kenton2019bert, brown2020language, touvron2023llama, touvron2023llama2, jiang2024mixtral, bai2023qwen}and large vision models~\cite{radford2021learning, kirillov2023segment, ravi2024sam, zhang2023prompt, zhang2023learning}, Multimodal Large Language Models (MLLMs) are rapidly becoming one of the hottest research topics among the community. By leveraging the power of LLMs, MLLMs can tackle a much wider range of tasks enabling them to work on multiple modalities~\cite{li2022blip, alayrac2022flamingo, li2023mimic,lyu2023macaw, li2024llava, ren2024timechat, wu2024towards, shenoy2024lumos,guo2023point, hong20233d}. Among the rapidly evolving MLLMs, GPT-4v~\cite{gpt4v}, and Claude 3.7~\cite{reid2024gemini} stand out for their exceptional performance on multiple visual understanding and reasoning tasks. 

Despite the impressive performance of state-of-the-art MLLMs on various general tasks, their effectiveness on SVG-related tasks remains limited, as indicated by previous researches~\cite{zou2024vgbench,ji2025socraticchartcooperatingmultiple}. Zou \textit{et al.} ~\cite{zou2024vgbench} employed MLLMs to generate captions that only serve as inputs for text-to-SVG generation. Ji \textit{et al.}~\cite{ji2025socraticchartcooperatingmultiple} developed an agentic workflow to generate SVGs from scalar chart images. However, these methods employ a workflow approach that sugarcoats the capability of MLLM in handling SVG-related tasks. In contrast, our work aims to explore the potential of MLLMs to address a series of SVG-centric tasks in an end-to-end manner. 

\begin{figure*}[htp]
\centering
\includegraphics[width=1\textwidth]{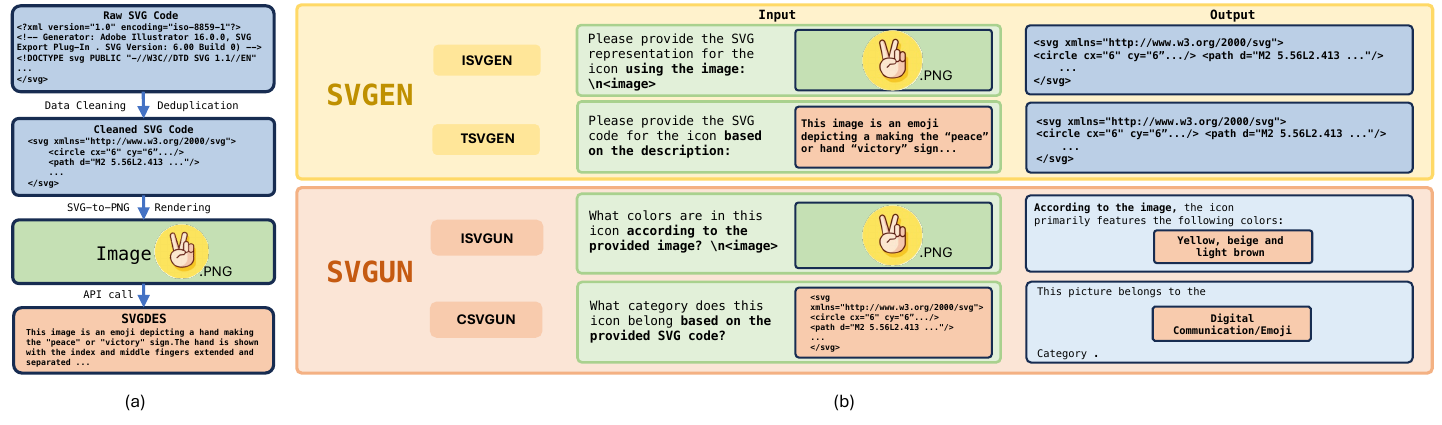}
\vspace{-0.2cm}
\caption{The Pipeline of UniSVG Construction. (a) illustrates the general UniSVG data pre-processing. (b) explains the construction process of each sub-dataset and provides an overview of the UniSVG dataset.}
\vspace{-0.1cm}
\label{figure2}
\end{figure*}

\section{UniSVG Dataset}
\label{sec:dataset}
In this section, we will introduce our SVG-centric dataset tailored for MLLM training--UniSVG, which is designed for unified SVG U$\&$G on different levels. Although SVGs are widely available on the internet, their inconsistent and noisy nature precludes their direct use for model training. Therefore, meticulous data cleaning and deduplication are essential to ensure data quality. In Section \ref{sec:Data Collection and Pre-processing}, we will discuss the detailed data collection and cleaning process. 

Upon obtaining cleaned SVG data, to fully unlock the potential of MLLMs solving SVG-related tasks following human instructions, we build a comprehensive SVG-centric multi-modal multi-task dataset. In Section \ref{sec:Dataset Construction}, we examine the dataset construction, including the transformation from SVG data into multiple modalities and the overall distribution of various tasks. In summary, UniSVG consists of three parts: image to SVG generation (ISVGEN), text to SVG generation (TSVGEN) and SVG understanding (SVGUN). The pipeline of the UniSVG dataset construction is shown in Figure \ref{figure2}. 

\subsection{Data Collection and Pre-processing}
\label{sec:Data Collection and Pre-processing}

SVG is easy to obtain and gather due to its widespread availability in code form on the internet and the numerous open-source resources. Conveniently, we collected 526k SVG codes from two online open-source SVG resources -- SVG icons\footnote{https://www.kaggle.com/datasets/victorcondino/svgicons} and SVGen-500k \footnote{https://huggingface.co/datasets/umuthopeyildirim/svgen-500k-instruct}. 
These datasets, sourced from publicly available repositories like SVG-Repo, offer a robust and diverse foundation of raw data for developing UniSVG.

To acquire raw SVG data that is cleaner and more suitable for MLLMs training, we have conducted a thorough and meticulous data cleaning and de-duplication process. This process encompasses several critical aspects, including:

{\setlength{\parindent}{0cm}\textbf{Simple SVG code cleaning.} We converted the SVG code in the original dataset to corresponding 336x336 PNG format images. Any SVG files that cannot be successfully converted subsequently removed from the dataset. Notably, we set the background color of all the SVG transformed images to be white.}

{\setlength{\parindent}{0cm}\textbf{Deep SVG code cleaning.} We optimized the SVG code by deleting the DOCTYPE and XML declarations in the header, comments, excising redundant nested \textless g\textgreater tags, and superfluous spaces.}

{\setlength{\parindent}{0cm}\textbf{SVG code deduplication.} We utilized perceptual hashing (pHash) \cite{Hadmi12} to encode PNGs rendered by SVGs, projected them into a representation space, and thereafter eliminated SVGs until the distances between each pair of SVGs fall below a specified threshold.}

{\setlength{\parindent}{0cm}\textbf{Elimination of SVGs with excessive Bezier curves.} When examining the dataset, we discovered several SVG-formatted manuscripts and sketches containing so many Bézier curves that they exceeded the model's modeling capabilities. To ensure model training stability, we excluded SVG files containing more than 100 Bézier curves.}

Following the comprehensive data cleaning and deduplication process, we obtained a collection of 360k vector graphics in clean SVG format, which constitute the foundation of our UniSVG dataset.

\begin{table}[t]
\belowrulesep=0pt
\aboverulesep=0pt
\centering
\small
\caption{Task-wise statistics of UniSVG Dataset. }
\vspace{-0.1cm}
\begin{tabular}{c|c|cc|cc}
\toprule
\centering
\multirow{2}{*}{\textbf{Dataset}} & \multirow{2}{*}{\textbf{Total}} &  \multicolumn{2}{|c|}{\textbf{SVGEN}}  & \multicolumn{2}{c}{\multirow{2}{*}{\textbf{SVGUN}}} \\
\cline{3-4}
&&ISVGEN&TSVGEN& & \\
\midrule
\multirow{2}{*}{Train} & \multirow{2}{*}{525,741} & 359,649 & 90,993 & \multicolumn{2}{c}{\multirow{2}{*}{75,099}}  \\
\cline{3-4}
& & \multicolumn{2}{c|}{450,642} & & \\

\midrule

\multirow{2}{*}{Train} & \multirow{2}{*}{2,850} & 1,000 & 500 & \multicolumn{2}{c}{\multirow{2}{*}{1,350}}  \\
\cline{3-4}
& & \multicolumn{2}{c|}{1,500} & & \\

\bottomrule
\end{tabular}
\label{table2}
\vspace{-0.1cm}
\end{table}

\subsection{Dataset Construction}
\label{sec:Dataset Construction}

In this section, we detail the construction process of UniSVG. In this process, by utilizing rendering techniques and leverage LLMs, we achieve full modality integration including description text, structural codes, and scalar images. As shown in Figure \ref{figure2}, UniSVG consists of two main components: SVG generation (SVGEN) and SVG understanding (SVGUN), both of which are combination of these three modalities mentioned before.

To be more specific, during the construction process, we utilized rendering techniques and GPT-4V for each data entry with a prompt to generate detailed descriptions of the SVGs, which we named them \textit{``SVGDES''}. Our prompt was designed to ensure that GPT-4V generated content in four parts: ``Overall Description'', ``Colors'', ``Categories'', and potential real-life ``Usage''. Subsequently, the construction of our dataset was based on extracting detailed information from these descriptions using pre-defined rules. 

In determining the data proportion for various subdatasets, we considered the differing difficulty levels of these SVG-related tasks. As a result, we increased the proportion of SVGEN data. The distribution ratios for SVGEN and SVGUN were adjusted to approximately 6:1 respectively, which we did ablation experiments in Appendix to determine the best ratio between the two parts of data.

Since \textit{``SVGDES''} is directly labeled by GPT-4V, it would be inappropriate to evaluate GPT-4V based on these data entries from SVGUN. Therefore, we have excluded these data entries from the UniSVG benchmark to ensure a fair and unbiased assessment of GPT-4V's performance. Nonetheless, this portion of the data entries is still included in the training set to enhance the MLLM's comprehension of image content. The basic distribution of all the data contained in the UniSVG dataset is shown in Table \ref{table2}. The detailed dataset construction of ISVGEN, TSVGEN and SVGUN will be introduced below:

{\textbf{Image to SVG Generation (ISVGEN)} is built on 360K cleaned SVG data, specifically designed to teach the model in generating SVG-formatted vector graphics from scalar images. Afterward, we rendered these SVGs to get 360k image-SVG pairs. To avoid restricted data pattern, we created 20 questions with the intent of ``Generate SVG code for this image'', and randomly sampled each image with a question. For training large multimodal models, ISVGEN uses image and the assigned requests as input prompts, with the corresponding SVG code as the targeted output. Subsequently, we sampled 1,000 data points to establish a test set as part of the UniSVG benchmark.}

\renewcommand{\arraystretch}{1.2}

\begin{table}[t]
\belowrulesep=0pt
\aboverulesep=0pt
\centering
\caption{Detailed Formation of SVGUN in UniSVG.}
\vspace{-0.1cm}
\scalebox{0.75}{
\begin{tabular}{c|c|c|cccc}
\toprule
\multirow{2}{*}{\textbf{Evaluation}} & \multirow{2}{*}{\textbf{Question}} & \multirow{3}{*}{\textbf{Tasks}}  & \multicolumn{4}{c}{\textbf{SVGUN}} \\
\cline{4-7}
\textbf{Metrics} & \textbf{Levels} & & \multicolumn{2}{c}{CSVGUN} & \multicolumn{2}{c}{ISVGUN} \\
\cline{4-7}
& & & Train & Test & Train & Test \\
\hline
\multirow{5}{*}{Accuracy} & \multirow{4}{*}{Easy} & size & 2K & 50 & -- & -- \\
& & color & 20K & 50 & 20K & 50 \\
& & shape\_count & 9K & 50 & -- & -- \\
& & transform\_count & 9K & 50 & -- & -- \\
\cline{2-7}
& \multirow{3}{*}{Middle} & category & 20K & 50 & 20K & 50 \\
\cline{1-1}
\multirow{4}{*}{Bert} & & rect\_description &  3K & 100 & -- & -- \\
& & circle\_description & 7K & 100 & -- & -- \\
\cline{2-7}
& \multirow{2}{*}{Hard} & usage & 60K & 200 & 60K & 200 \\
& & general\_description & 60K & 200 & 60K & 200 \\
\hline
\multicolumn{3}{c|}{\multirow{2}{*}{\textbf{Total}}} & 190K & 850 & 160K & 500 \\
\cline{4-7}
\multicolumn{3}{c|}{} &  \multicolumn{4}{c}{\textbf{351,350}} \\
\bottomrule
\end{tabular}
}
\label{table3}
\vspace{-0.3cm}
\end{table}
\renewcommand{\arraystretch}{1}

{\textbf{Text to SVG Generation (TSVGEN)} contains 91k text-SVG pairs, specifically designed to help the model generate SVG-formatted vector graphics from descriptive texts. Considering huge cost of GPT-4V API call, we randomly selected a smaller set of 91k SVG data entries from the cleaned dataset to serve as the foundational data for TSVGEN. Similarly to \textbf{ISVGEN}, ``SVGDES” together with its corresponding SVG codes consist of text-SVG pairs. We also devised 20 queries to serve as instructs and randomly paired each text-SVG pair with one query. When utilized as training data for MLLMs, TSVGEN employs randomly paired ``questions'' and detailed descriptions as input, with the SVG code serving as the output. Notably, the corresponding images are not presented to the model during the training process. As a result, we sampled 1,000 data points from TSVGEN to form a test set, which is included in the UniSVG benchmark.}

{\textbf{SVG Understanding (SVGUN)} focused on helping the MLLMs understand SVG formatted image from different levels and different perspectives. As shown in Table \ref{table3}, in order to test MLLMs’ ability of understanding SVG, we divided \textbf{SVGUN} into three levels based on the difficulty of the tasks: Easy, Middle and Hard. The \textit{Easy} level includes tasks extracting four things: original size shown in the SVG codes, colors shown in the SVG images, quantity of different shapes used in SVG codes, and quantity of different transforms of $<$path$>$ used in SVG codes. The \textit{Middle} level includes tasks extracting three things: category of the SVG image, description of the rectangles in the SVG codes and the description of the circles in the SVG codes. The \textit{Hard} level includes two main tasks: general description of the entire SVG image and speculating the possible usage of the SVG image in the real life. Notably, \textbf{SVGUN} contains two sub-tasks categorized by reference modality: Image-based SVG Understanding (ISVGUN) and Code-based SVG Understanding (CSVGUN), which differs from the format of the input: CSVGUN uses the SVG code and the question as the input, while ISVGUN uses image and the question as the input. }

\begin{table*}[t]
\aboverulesep=0pt
\belowrulesep=0pt
\centering
\caption{Main Results of our experiments. Finetuned + MODEL means that we adopted pretrained weights of the model and finetuned the model on UniSVG dataset. We highlight the best performance for each evaluation metric in bold.}
\vspace{-0.15cm}
\scalebox{0.77}{
\begin{tabular}{c|c|cccc|cccc|cc|cc}
\toprule
\multirow{4}{*}{\textbf{Model Name}} &\multirow{4}{*}{\textbf{Final Score}} & \multicolumn{8}{c|}{\textbf{SVGEN}} &  \multicolumn{4}{c}{\textbf{SVGUN}}  \\ 
\cline{3-14}
& &\multicolumn{4}{c|}{ISVGEN} & \multicolumn{4}{c|}{TSVGEN}  & \multicolumn{2}{c|}{Accuracy} & \multicolumn{2}{c}{Bert}\\ 
& &\multicolumn{2}{c}{Low-Level} & High-Level & \multirow{2}{*}{Score} &\multicolumn{2}{c}{Low-Level}& High-level & \multirow{2}{*}{Score} & \multirow{2}{*}{Easy-Acc $\uparrow$} & \multirow{2}{*}{Hard-Acc$\uparrow$ } & \multirow{2}{*}{Bert $\uparrow$} & \multirow{2}{*}{SBert $\uparrow$}\\  
& & SSIM $\uparrow$ & LPIPS $\downarrow$ & CLIP Score $\uparrow$ & & SSIM $\uparrow$ & LPIPS $\downarrow$ & CLIP Score $\uparrow$ & &  &  &  \multicolumn{2}{|c}{}   \\ 
\midrule
LLaVA 1.5 13B & 0.569 & 0.547& 0.683 & 0.711 & 0.599 & 0.574 & 0.640 & 0.729 & 0.624 & 0.321 & 0.224 & -0.154 & 0.473\\
LLaMA 3.2 & 0.567 & 0.563 & 0.674 & 0.690 & 0.592 & 0.491 & 0.616 & 0.772 & 0.638 & 0.347 & 0.201 & -0.291 & 0.455 \\
Qwen 2.5 VL & 0.614 & 0.564 & 0.614 & 0.738 & 0.633 & 0.538 & 0.619 & 0.764 & 0.642 & 0.543 & 0.571 & 0.082 & 0.596  \\
Gemini 1.5 pro & 0.647 & 0.538 & 0.608 & 0.764 & 0.644 & 0.572 & 0.558 & 0.798 & 0.681 & 0.923 & 0.523 & 0.254 & 0.637\\
GPT 4V & 0.650 & 0.557 & 0.582 & 0.740 & 0.638 & 0.620 & 0.531 & 0.816 & 0.712 & 0.893 & 0.448 & 0.211 & 0.520\\
Claude 3.7 & 0.722 & 0.622 & 0.473 & \textbf{0.855} & 0.743 & \textbf{0.622} & 0.503 & 0.852 & 0.735 & 0.863 & \textbf{0.624} & 0.322 & 0.709 \\
\midrule
Finetuned LLaVA 1.5 & 0.689 & 0.654 & 0.479 & 0.802 & 0.716 & 0.586 & 0.544 & 0.787 & 0.680 & 0.883 & 0.521 & 0.473 & 0.728 \\
Finetuned LLaVA-LLaMA & 0.731 & 0.713 & 0.402 & 0.837 & 0.764 & 0.617 & 0.503 & 0.823 & 0.717 & 0.940 & 0.556 & 0.524 & 0.759\\
Finetuned LLaMA 3.2 & 0.732& 0.722 & 0.378 & 0.843 & 0.775 & 0.617 & 0.511 & 0.812 & 0.709 & 0.990 & 0.560 & 0.511 & 0.735\\
Finetuned Qwen 2.5 VL & \textbf{0.752} & \textbf{0.725} & \textbf{0.368} & 0.836 & \textbf{0.773} & 0.621 & \textbf{0.479} & \textbf{0.853} & \textbf{0.740} & \textbf{0.983} & 0.604& \textbf{0.574} & \textbf{0.827} \\
\bottomrule
\end{tabular}
}
\vspace{-0.1cm}
\label{table4}
\end{table*}

\section{Experiments}
In the experimental section, we will present UniSVG-Bench, the results of fine-tuning open-source models, additional ablation studies, and visualizations.\label{sec:experiments}

\subsection{Benchmark and Metrics}
To thoroughly test the understanding and generation capabilities of AI in SVG-related tasks, we constructed the UniSVG benchmark by selecting a test set of 2,850 data items from the UniSVG dataset. Furthermore, we employed diverse and multi-dimensional evaluation metrics to rigorously assess model performance across various tasks, ensuring a comprehensive evaluation.

\textbf{Metrics for SVGEN:} We compared the similarity between the images rendered from generated SVG code those rendered from the corresponding ground-truth SVG codes. 
 On one hand, we utilized Structural Similarity Index (\textbf{SSIM}) \cite{1284395} and Learned Perceptual Image Patch Similarity (\textbf{LPIPS}) \cite{zhang2018unreasonableeffectivenessdeepfeatures} as evaluation metrics, both of which use local comparison methods to evaluate image similarity, focusing on local regions to catch low-level visual features. On the other hand, we employed \textbf{CLIP similarity} \cite{radford2021learning} as the evaluation metric, which differs from SSIM and LPIPS by leveraging pre-trained VL models to capture semantic similarities, offering semantic-aware and high-level understanding of image content.
 
\textbf{Metrics for SVGUN: }As shown in Table \ref{table3}, we used \textbf{Accuracy} to compare the model's output with the ground-truth answers for simpler tasks. For more challenging tasks and freely generated model outputs, we assessed the similarity between the model-generated responses and the ground-truth sentences using two popular model-based metrics: \textbf{BERTScore} \cite{zhang2020bertscoreevaluatingtextgeneration} and Sentence-BERT (\textbf{SBERT}) \cite{reimers2019sentencebertsentenceembeddingsusing}.

To rank the models on the leaderboard, a final score should be derived by calculating a weighted sum of all evaluation metrics, thereby measuring their overall capabilities. In the UniSVG-bench scoring system, we have allocated 45$\%$ each to ISVGEN and TSVGEN, considering their higher difficulty and significance compared to SVGUN. Consequently, SVGUN accounts for the remaining 10$\%$ of the total score. From a more detailed perspective, within the SVGEN evaluation metrics, we have assigned a higher weight to CLIP similarity due to its greater importance. Specifically, CLIP similarity accounts for 60$\%$ of the score within both ISVGEN and TSVGEN, while SSIM and LPIPS each account for 20$\%$. Similarly, the metrics within SVGUN have been adjusted according to the difficulty of the tasks, with the proportions for easy accuracy, hard accuracy, BERT, and SBERT set at 1:2:3.5:3.5.

\subsection{Main Results}

To verify the effectiveness of the UniSVG dataset in enhancing the performance of MLLMs for SVG U$\&$ tasks, we selected a variety of popular open-source MLLMs for fine-tuning. As shown in Table \ref{table4}, after fine-tuning, multiple MLLMs have matched or even surpassed the performance of SOTA closed-source models.

Specifically, after fine-tuning on UniSVG for 3 epochs, Qwen 2.5 VL achieved best overall performance among all models including fine-tuned and close-source models. Claude 3.7 achieved the best performance among all the closed-source SOTA models. Notably, the CLIP similarity of Claude 3.7 is the highest compared to all other models. However, the pixel-level evaluation metrics are pretty low compared with finetuned MLLMs, which indicates that the generated SVG image of Claude 3.7 is semantically similar to the target image but lacks visual detail and structural accuracy. Additionally, the LLaVA-LLaMA model, using LLaMA 3.1 as its LLM, showed impressive performance compared to vanilla LLaVA 1.5 (Vicuna 1.5 as its LLM). This indicates that integrating a more advanced LLM, especially one proficient in code generation, can enhance SVG-centric task performance within the same MLLM framework.

In summary, our experiments demonstrate that fine-tuning on the UniSVG dataset significantly enhances the performance of MLLMs for SVG generation and understanding tasks.

\subsection{Ablation Study}
\label{ablation}

In this section, we conducted a series of ablation experiments aiming to answer the following research questions. Additionally, data ratio and data scaling experiments are detailed in the Appendix.

\begin{itemize} 
    \item \textbf{RQ1} What is the most effective fine-tuning paradigm for MLLMs on SVG-centric tasks? 
    \item \textbf{RQ2} Given the unique characteristics of SVG codes, are there strategies to improve training efficiency? 
\end{itemize}

\subsubsection{Finetuning Experiments on LLaVA (RQ1)}

We conducted various fine-tuning experiments on the LLaVA and LLaVA-LLaMA models to identify the best fine-tuning paradigm. The results are presented in Table \ref{table5}.

LLaVA, like most MLLMs, uses a two-stage training paradigm: pre-training (stage 1: aligning visual and language) and SFT (stage 2: training projector and MLLM to follow instructions). Our experiments showed that fine-tuning the stage 1 model on the UniSVG dataset led to superior performance in all SVG-related metrics. This suggests that stage 1 training is beneficial, while stage 2 training may be detrimental for SVG post-training.

Similar experiments are conducted on LLaVA-LLaMA and fine-tuning the stage 1 model on UniSVG achieves the best performance.

\begin{figure*}[t]
\centering
\includegraphics[width=\textwidth]{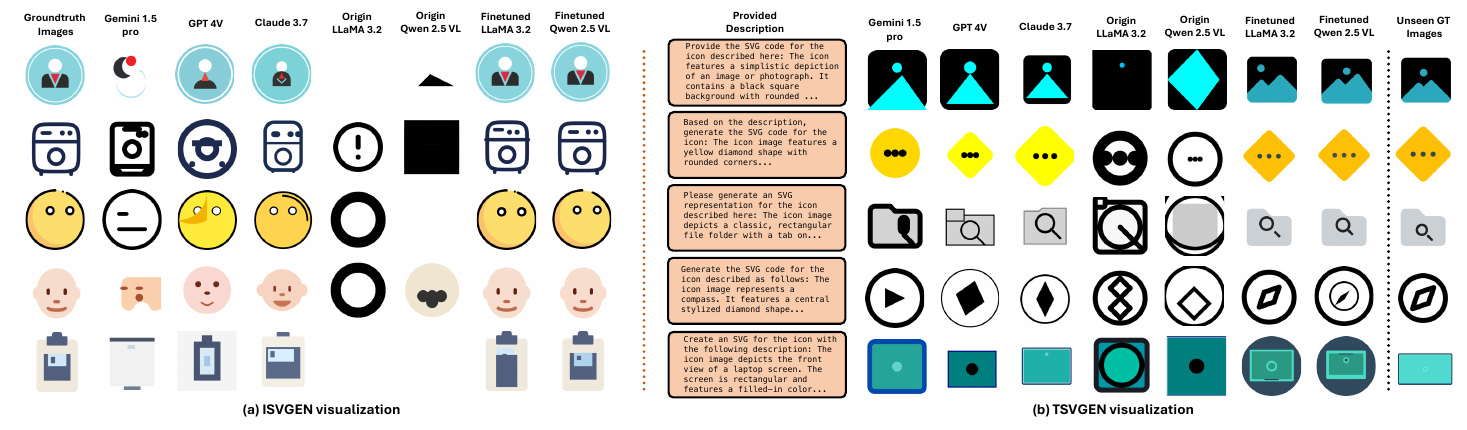}
\vspace{-0.3cm}
\caption{Visualization of SVGs generated by MLLMs conditioned on images (left) and texts (right). Note that ``Unseen GT image'' is not used during the generation from textual description.}
\label{figure3}
\end{figure*}

\renewcommand{\arraystretch}{1.2}
\begin{table}[!t]
\belowrulesep=0pt
\aboverulesep=0pt
\centering
\caption{Performance comparison with different finetune methods on LLaVA 1.5 and LLaVA-LLaMA}
\vspace{-0.15cm}
\scalebox{0.70}{
\begin{tabular}{c|ccc|c}
\toprule
\multirow{2}{*}{\textbf{Model Name}} & \multicolumn{3}{c|}{\textbf{Finetune Stages}} & \multirow{2}{*}{\textbf{Final Score}} \\
& \textbf{Stage 1} & \textbf{Stage 2} & \textbf{UniSVG} & \\
\midrule
\multirow{3}{*}{LLaVA 1.5} & $\times$ & $\times$ & $\checkmark$ & 0.665 \\
& $\checkmark$ & $\checkmark$ & $\checkmark$ & 0.672 \\
& $\checkmark$ & $\times$ & $\checkmark$ & \textbf{0.689} \\
\midrule
\multirow{2}{*}{LLaVA-LLaMA} & $\times$ & $\times$ & $\checkmark$ & 0.716 \\
& $\checkmark$ & $\times$ & $\checkmark$ & \textbf{0.731} \\
\bottomrule
\end{tabular}
}
\vspace{-0.2cm}
\label{table5}
\end{table}
\renewcommand{\arraystretch}{1}

\subsubsection{SVG-specific Optimization of MLLM Training Efficiency (RQ2)}
SVG format images contain numerous redundant commas and decimal points that significantly impede training efficiency. Our ablation study on the ISVGEN task showed that removing them reduced training tokens by 35$\%$, as shown in Table~\ref{table6}. Although this optimization slightly decreased SVG generation quality due to the model's adaptation to the modified SVG representation, it provides valuable insights for future MLLM training with SVG data, which appeals for balancing computational efficiency with model performance.

\renewcommand{\arraystretch}{1.2}
\begin{table}[!t]
\belowrulesep=0pt
\aboverulesep=0pt
\centering
\caption{Training efficiency and performance comparison before and after removing redundant tokens.}
\vspace{-0.15cm}
\scalebox{0.6}{
\begin{tabular}{l|cc|cc|cccc}
\toprule
\multirow{2}{*}{\textbf{Model}} & \multicolumn{2}{c|}{\textbf{Token Reduction}} & \multicolumn{2}{c|}{\textbf{Time Reduction}} & \multicolumn{4}{c}{\textbf{ISVGEN Performance}} \\ \cline{2-9}
& \textbf{Avg. Num} & \textbf{Ratio} & \textbf{Avg. Time} & \textbf{Ratio} & \textbf{SSIM} & \textbf{LPIPS} & \textbf{CLIP Score} & \textbf{Total} \\
\midrule
LLaMA 3.2 (Before) & 1,077 & \multirow{2}{*}{35.6$\%$} & 29.6h & \multirow{2}{*}{39.5$\%$} & 0.673 & 0.501 & 0.776 & 0.700 \\
LLaMA 3.2 (After) & 694 &  & 17.9h &  & 0.598 & 0.583 & 0.748 & 0.651 \\ \midrule
GLM 4V (Before) & 1,214 & \multirow{2}{*}{37.1$\%$} & 31.6h & \multirow{2}{*}{42.7$\%$} & 0.613 & 0.533 & 0.731 & 0.655 \\ 
GLM 4V (After) & 764 &  & 18.1h &  & 0.601 & 0.578 & 0.712 & 0.632 \\
\bottomrule
\end{tabular}
}
\vspace{-0.2cm}
\label{table6}
\end{table}
\renewcommand{\arraystretch}{1}

\subsection{Visualization and Demonstration}
\label{visual}

In this section, we present a comprehensive visualization to showcase the enhanced SVG generation capabilities of the fine-tuned MLLMs, compared with SOTA close-source models.

As illustrated in Figure \ref{figure3}, fine-tuning on the UniSVG dataset significantly enhances the SVG generation capabilities of the model. Some details even surpass those of SOTA proprietary models. Notably, Qwen 2.5VL excels in SVGEN tasks, while Claude 3.7 remains the top proprietary model, consistent with the data in Table \ref{table4}.

\begin{figure}[t]
\centering
\includegraphics[width=0.5\textwidth]{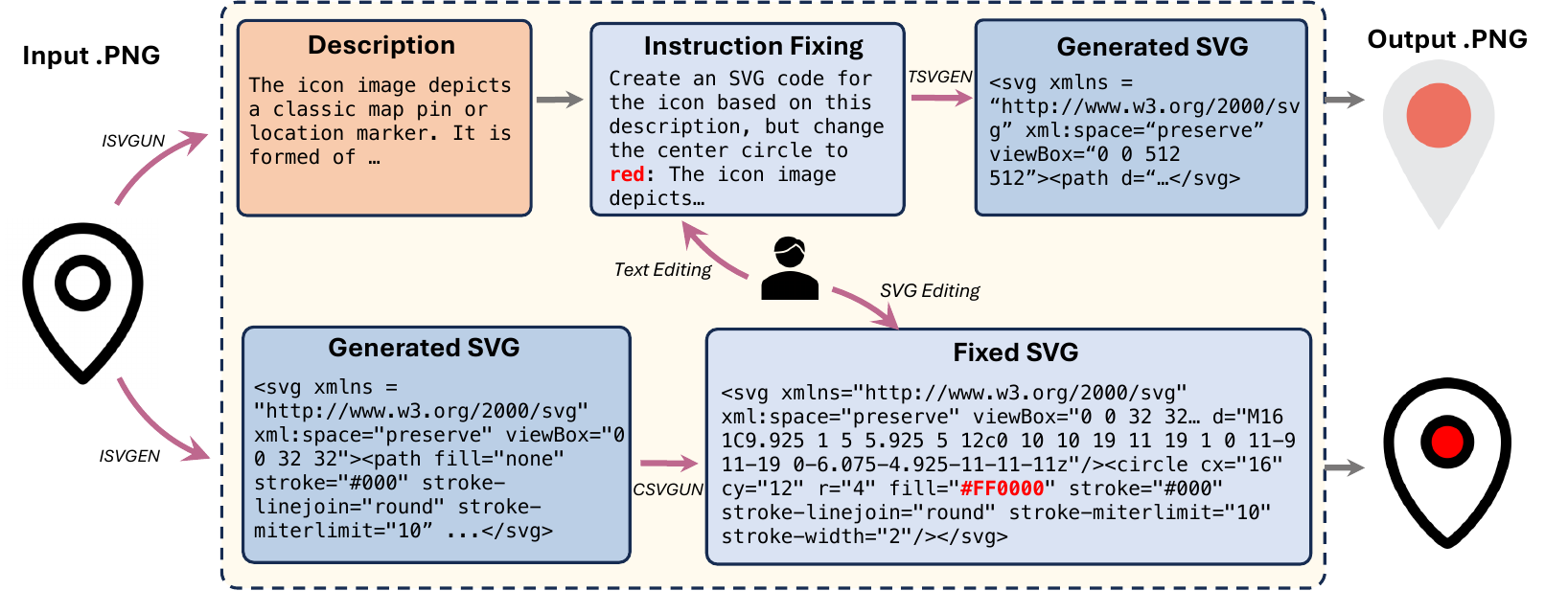}
\caption{Human-in-the-loop SVG-based image editing using fine-tuned Qwen 2.5 VL}
\label{figure4}
\vspace{-0.2cm}
\end{figure}

In addition, we built an SVG-based image-editing demo to showcase the application of fine-tuned Qwen 2.5 VL in a human-in-the-loop manner. As illustrated in Figure \ref{figure4}, a human user have two pathways to achieve SVG-based image-editing: (1) edit the model-generated description and then produce a new SVG from the texts, or (2) generate SVG code from the given image and modify the code to create the desired image. The straightforward demo highlights the potential for building powerful interactive workflow with the fine-tuned unified models for SVG-related tasks.
\section{Conclusion and Future Work}
\label{sec:conclusion}

In this work, we construct the UniSVG to facilitate SVG generation and understanding in the era of MLLMs. The benchmark includes 528k data items and spans a variety of downstream tasks including Image2SVG, Text2SVG and SVG Understanding, making it a comprehensive resource for the field. Our experiments on this dataset demonstrate that fine-tuning on UniSVG significantly outperforms state-of-the-art models on multiple SVG-related tasks, including closed-source models like GPT-4V and Claude 3.7.

Looking ahead, our fine-tuning experiments indicate significant potential for MLLM improvement in SVG U$\&$G. We hope that UniSVG will be widely adopted to accelerate SVG-based research. More fundamentally, compared to bitmap images, SVG offers higher abstraction and greater information density, and breakthroughs in SVG U$\&$G could have far-reaching implications for MLLM development with SVG as the central visual modality.

\newpage
\section*{ACKNOWLEDGMENTS}
\label{ACKNOW}
We would like to express our sincere gratitude to Yi Liu and Feng Liu for their invaluable support and resources provided throughout this research.

\bibliographystyle{ACM-Reference-Format}
\balance
\bibliography{sample-base}

\end{document}